# 3D CNN with Localized Residual Connections for Hyperspectral Image Classification

Shivangi Dwivedi[1], Murari Mandal[2], Shekhar Yadav[1] and Santosh Kumar Vipparthi[2*]

[1]Madan Mohan Malaviya University of Technology, Gorakhpur
[2] Vision Intelligence Lab, Malaviya National Institute of Technology Jaipur
`shivangid48@gmail.com, murarimandal.cv@gmail.com,`
`syee@mmmut.ac.in, *skvipparthi@mnit.ac.in`

**Abstract.** In this paper we propose a novel 3D CNN network with localized residual connections for hyperspectral image classification. Our work chalks a comparative study with the existing methods employed for abstracting deeper features and propose a model which incorporates residual features from multiple stages in the network. The proposed architecture processes individual spatio-spectral feature rich cubes from hyperspectral images through 3D convolutional layers. The residual connections result in improved performance due to assimilation of both low-level and high-level features. We conduct experiments over Pavia University and Pavia Center dataset for performance analysis. We compare our method with two recent state-of-the-art methods for hyperspectral image classification method. The proposed network outperforms the existing approaches by a good margin.

## 1    Introduction

Advancements in sensing technologies have led to improved spectral resolution of hyperspectral images. Digitization and automatic analysis of hyperspectral images have enabled exploration of remote areas in an efficient way. It has enabled the possibility of forecasting geographical events before their occurrence and to map the effects of any geographical phenomena. Moreover. hyperspectral images also find application in various fields like biological threat detection, fire-tracking problem, landcover usage applications like agriculture, construction, etc. [1, 2]. However, automatic analysis of these images such as pixel-wise classification is still a challenging task. Existing methods designed for classification of spatio-spectral images lack the desired efficacy. Thus, special modifications are required in existing methods to improve the quality of information abstraction from these images.

Satellite images are captured by different type of sensors (Aviris, Enmap, Hyperion, etc.) in different spectral formats. The spectral formats can be subdivided into 3 types based on the spectral information they carry, namely, panchromatic, multi-spectral and hyperspectral. All three can have same spatial information but vary in their respective spectral information. Panchromatic images are gray scale images whereas multispectral and hyperspectral images contain tens or hundreds of spectral bands respectively. These spectral bands range from continuous and narrow (for hyperspectral images) to discrete and wider bands for a pixel (for multi-spectral images).

Analyzing classification accuracy of hyperspectral images is the main concern of this paper. Hyperspectral images captured by sensors are rich in spatio-spectral information but suffer from curse of dimensionality as well. Therefore, the conventional methods have used pre-processing [3-7] before classifying these images. It involves removal of unwanted, irrelevant bands from each pixel of the image. The existing approaches for hyperspectral image classification (HSIC) can be grouped into traditional [3-7, 16, 17] and deep learning-based methods [18-35]. Deep learning models have been much more effective in HSIC as compared to the traditional approaches. We investigated some recent CNN models for HSIC and identified some of the network design enhancements to improve classification accuracy.

In this paper we proposed a new 3D-CNN network with localized residual connections for HIC. The proposed network learns abstract representation from raw spectral signals at each pixel location through gradual reduction in number of 3D convolutional kernels. Moreover, at multiple convolution stages, we add residual features from previous layers to reinforce the low-level features. The spatio-spectral features learned from both low-level and high-level abstractions led to improved performance over both PaviaU and PaviaC datasets.

## 2 Related Work

Hyperspectral (HS) images comprise of a contiguous, narrow spectrum of informational bands. Hence feature extraction and classification of regions in a scene is a very challenging task. In [4], each pixel is smoothened by a weighted mean filter. Then spectral regularized scatter matrix is integrated with spatial neighborhood scatter matrix to find local similarity pattern for classification. The authors in [5] used mahalanobis distance metric to calculate intra-class and inter-class distance between pixels. Koonsanit et al. [6] use combination of PCA and integrated gain to select relevant band. Similarly, Li et al. [7] used Fisher's discriminant analysis to get rid of redundant bands for hyperspectral image analysis. Spatial filtering technique [16] is one of the conventional methods in which image is divided into its constituent spatial frequency. Then selected altering of certain frequency is done to suppress some frequencies and emphasize other features. Moreover, PCA features with SVM classifiers are used in [17].

Recently, many deep learning models have also been designed for hyperspectral image classification. We give a brief description of different model architectures proposed by various researchers. Li et al. [8] proposed a simple 3D CNN whereas, Chen et al. [9] added L2 regularization and dropouts to address the problem of overfitting. Mei et al. [10] enhanced the HS image by designing a network for spatial super resolution. Liu et al. [11] utilized bidirectional recurrent connections to abstract minute details form the image. Zhu et al. [12] presented a generative adversarial network comprised of two parts 1D and 2D convolutions. He et al. [13] used five layered multi-scale 3D CNN and Chen et al. [14] used logistic regression as final classifier for combined data of LIDAR and HS image.

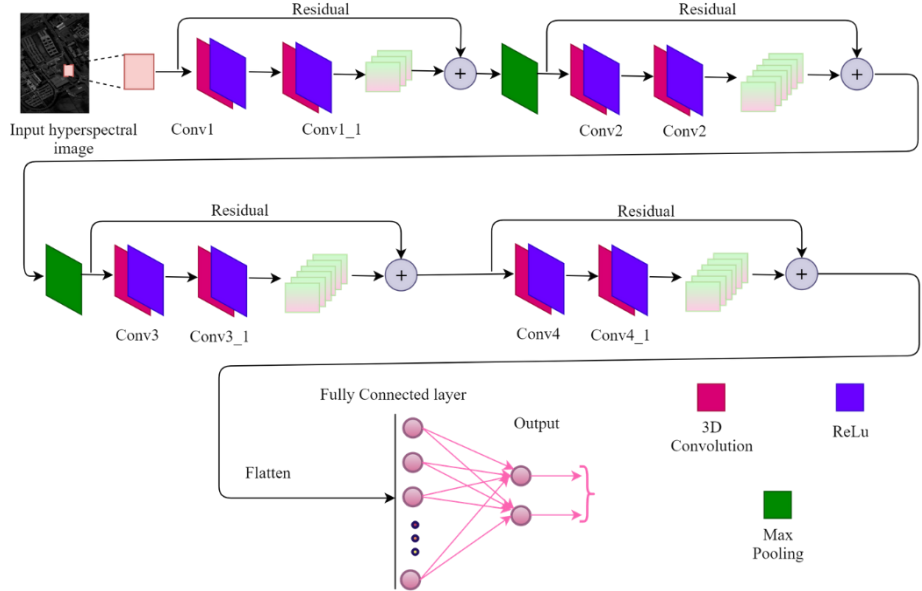

**Fig. 1.** The architecture of the proposed 3D CNN with localized residual connections.

Lee et al. [18] proposed a contextual deep CNN by concurrently extracting multiple 3-dimensional local convolutional features with different sizes jointly exploiting spatial and spectral features of HS image. Hamida et al. [19] proposed a 3D CNN prototype to facilitate joint study of spatial and spectral features of HS image. Similarly, many other CNN models have been designed in the literature for HSIC and aerial images analysis [20-25].

## 3 Proposed Method

This paper introduces a novel 3D CNN architecture using residual blocks for HSIC. Residual blocks are used to retain the features from earlier convolutional layers. We use four residual blocks to assimilate both primitive and abstract information. Similarly, the spatio-spectral are features learned from both low-level and high-level abstractions leading to robust performance. We depict the proposed 3D CNN architecture in Fig. 1. Each pixel in a hyperspectral (HS) image can be seen as a cube containing spatial and spectral data. Each 3-dimensional cube is configured into 1×1×S structure. Of this 1×1 is spatial data dimension and S is taken as continuous, narrow spectrum bands. These spectrum bands are digitally represented with multiple channels in HS images.

Let $HSI_S(a,b)$ be an input HS image of size P×Q×S, $a \in [1,P]$, $b \in [1,Q]$ having $S$ channels. From $S$ spectral channels, the 3D convolutional features (3DCF) are computed using Eq. (1) - Eq. (5).

$$3DCF(HSI_S) = \psi_4(\psi_3(\psi_2(\psi_1(HSI_S)))) \tag{1}$$

$$\psi_1(z) = ap_{1,1,3}(\kappa_{20,1,1,1}(\Re(\kappa_{20,3,3,3} \otimes z))) \tag{2}$$

**Table 1.** Detailed network architecture of the proposed network

| Layer Name | Kernel | Stride | Inputs | Outputs |
|---|---|---|---|---|
| Conv1 | 3x3x3 | 1, 1, 1 | 102 | 20 |
| Conv1_1 | 1x1x1 | 1, 1, 1 | 20 | 20 |
| Pool1 | 1x1x3 | 1, 1, 2 | 20 | 20 |
| Conv2 | 3x3x3 | 1, 1, 1 | 20 | 35 |
| Conv2_1 | 1x1x1 | 1, 1, 1 | 35 | 35 |
| Pool2 | 1x1x3 | 1, 1, 2 | 35 | 35 |
| Conv3 | 1x1x3 | 1, 1, 1 | 35 | 35 |
| Cov3_1 | 1x1x1 | 1, 1, 1 | 35 | 35 |
| Conv4 | 1x1x2 | 1, 1, 2 | 35 | 35 |
| Conv4_1 | 1x1x1 | 1, 1, 1 | 35 | 35 |

$$\psi_2(z) = ap_{1,1,3}(\kappa_{30,1,1,1}(\mathfrak{R}(\kappa_{35,3,3,3} \otimes z))) \tag{3}$$

$$\psi_3(z) = \kappa_{35,1,1,1}(\mathfrak{R}(\kappa_{35,1,1,3} \otimes z)) \tag{4}$$

$$\psi_4(z) = \kappa_{35,1,1,1}(\mathfrak{R}(\kappa_{35,1,1,2} \otimes z)) \tag{5}$$

where $\kappa_{n,h,w,d}$ denotes 3D convolution with parameters $n$, $h$, $w$ and $d$ representing the number of kernels, height, width and depth of the kernels respectively. We use strides (0,0,0), (0,0,0) and (0,0,1) in three layers of $\psi_1$ and $\psi_2$ blocks. Similarly, we use strides (0,0,1) and (0,0,0) in two layers of $\psi_3$ $\psi_4$ blocks. The $ap_{h,w,d}$ represents 3D average pooling and $\mathfrak{R}(\cdot)$ denotes the rectified linear unit (ReLu) activation function.

The proposed model convolves input HS image with filter *Conv1* of shape of 3×3×3 taking input channel as 102 or 103 and output channel as 20. Here, spatial window of 7×7 is selected for feature extraction at each pixel. Then further convolving it with filter *Conv1_1* (1×1×1), taking input and output channels to be 20. The identical value of input is then added to output of the second convolving layer after applying ReLu non-linearity. In short, residue between input to first convolving layer and output of second convolving layer is calculated. The pooling layer then comes in function reducing spectral depth of the image with filter size of 1×1×3 at stride of 2 along depth, 1 along height and width; adding zero padding only across depth.

Similar convolution layers with residual block is replicated as shown in Fig. 1 and Eq. (1) – Eq. (5). The final stack of convolution blocks (*Conv4*, *Conv4_1*) applies the localized residuals as follows. *Conv4* has filter of size 1×1×2 at stride of 1 along height, width and of 2 along depth of the image. Zero padding is also added along depth of the image. The *Conv4_1* convolution involves use of filter of size of 1×1×1 at stride of 1 along all the directions. This is further connected to a fully connected (FC) layer after flattening the output of convolution stack. The detailed network architecture of the proposed methods is tabulated in Table 1.

# 4    Experiments and Discussions

In this section, we conduct multiple experiments to demonstrate the effectiveness of the proposed residual based 3D CNN design. We first discuss about the HS image datasets used in our experiments. Moreover, the hyperparameters and other training configurations are also explained. Finally, we discuss experimental results of the proposed and existing state-of-the-art approaches in detail.

**Table 2.** Data division for model training in PaviaU dataset

| PaviaU Classes | Train samples | Test samples |
|----------------|---------------|--------------|
| Asphalt | 200 | 6431 |
| Meadows | 200 | 18449 |
| Gravel | 200 | 1899 |
| Trees | 200 | 2864 |
| Sheets | 200 | 1154 |
| Bare Soil | 200 | 4829 |
| Bitumen | 200 | 1130 |
| Bricks | 200 | 2482 |
| Shadows | 200 | 747 |

**Table 3.** Data division for model training in PaviaC dataset

| PaviaC Classes | Train samples | Test samples |
|----------------|---------------|--------------|
| Water | 200 | 624 |
| Trees | 200 | 620 |
| Asphalt | 200 | 616 |
| Self-Blocking Bricks | 200 | 608 |
| Bitumen | 200 | 608 |
| Tiles | 200 | 1060 |
| Shadows | 200 | 276 |
| Meadows | 200 | 624 |
| Bare Soils | 200 | 620 |

**Table 4.** Number of trainable parameters at each layer of the proposed model

| Network layers | # Parameters |
|----------------|--------------|
| Conv1 | 560 |
| Conv1_1 | 420 |
| Conv2 | 18,935 |
| Conv2_1 | 1,260 |
| Conv3 | 3,710 |
| Conv3_1 | 1,260 |
| Conv4 | 2,485 |
| Conv4_1 | 1,260 |
| Total | 29,890 |

### 4.1    Datasets

The proposed network has been trained and evaluated on two publicly available datasets Pavia University (PaviaU) and Pavia Center (PaviaC). PaviaU and PaviaC HS images were acquired by ROSIS sensor in a campaign over northern Italy. Each of these scenes comprise 1.3m spatial resolution. PaviaU is 610×340 resolution image and contains 103 band spectrums. Its ground truth differentiates the entire scene in 9 classes. PaviaC is 1096×1096 resolution image and consists of 102 band spectrums. Its ground truth groups the entire scene in 9 classes. Composed of rich spatial and spectral information, these datasets offer excellent platform for testing and evaluating performance of proposed network prototype. Training and testing sample size for each class in both the datasets is stated in Table 2 and Table 3.

### 4.2    Training Configuration

The proposed model is trained using stochastic gradient descent with momentum of 0.9 at learning rate of 0.02. Weight decay rate is set to 0.0005. We also experimented with Adam optimizer which did not produce good result. Zero padding is added at some layers to maintain input and output shape of the image while abstracting features of the input spatial-spectral rich image. Table 4 represents number of trainable parameters generated at each layer of the proposed model.

### 4.3    Results and Discussions

The proposed method has been trained and evaluated over PaviaC and PaviaU datasets. The performance is compared with recent state-of-the-art approaches [18, 19] in terms of classification accuracy. We trained our model with different percentage of training data (4.4%, 5%, 9%, 15%). The accuracy for different training sizes is depicted in Fig.

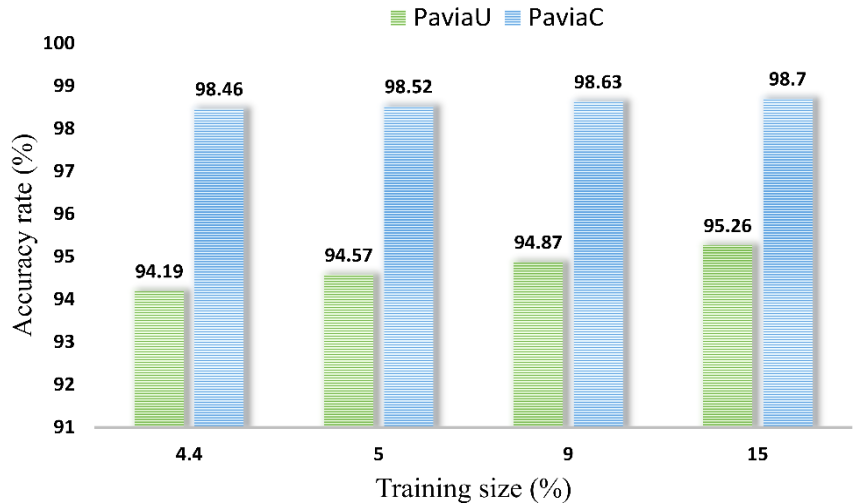

**Fig. 2.** The performance of the proposed model on PaviaU and PaviaC datasets with different percentages of training data.

| | *Predicted Label* | | | | | | | | |
|---|---|---|---|---|---|---|---|---|---|
| | **Asphalt** | **Meadows** | **Gravel** | **Trees** | **Sheets** | **Bare Soil** | **Bitumen** | **Bricks** | **Shadows** |
| **Asphalt** | 6041 | 7 | 15 | 0 | 1 | 0 | 21 | 70 | 0 |
| **Meadows** | 0 | 16457 | 0 | 6 | 0 | 20 | 0 | 0 | 0 |
| **Gravel** | 9 | 2 | 1660 | 0 | 0 | 0 | 0 | 260 | 0 |
| **Trees** | 0 | 70 | 0 | 2800 | 0 | 16 | 0 | 0 | 0 |
| **Sheets** | 0 | 0 | 0 | 0 | 1286 | 0 | 0 | 0 | 0 |
| **Bare Soil** | 6 | 32 | 0 | 0 | 0 | 4770 | 0 | 0 | 0 |
| **Bitumen** | 64 | 0 | 0 | 0 | 0 | 0 | 1202 | 2 | 3 |
| **Bricks** | 26 | 19 | 59 | 0 | 0 | 3 | 0 | 3413 | 0 |
| **Shadows** | 8 | 5 | 0 | 0 | 0 | 1 | 0 | 0 | 891 |

**Fig. 3.** Confusion matrix of the proposed method over PaviaU dataset

| | *Predicted Label* | | | | | | | | |
|---|---|---|---|---|---|---|---|---|---|
| | **Asphalt** | **Meadows** | **Gravel** | **Trees** | **Sheets** | **Bare Soil** | **Bitumen** | **Bricks** | **Shadows** |
| **Asphalt** | 62319 | 0 | 0 | 0 | 0 | 0 | 0 | 0 | 0 |
| **Meadows** | 0 | 6837 | 338 | 0 | 0 | 0 | 0 | 0 | 0 |
| **Gravel** | 0 | 52 | 2697 | 0 | 5 | 0 | 0 | 0 | 0 |
| **Trees** | 0 | 0 | 0 | 2553 | 14 | 0 | 0 | 0 | 0 |
| **Sheets** | 0 | 1 | 0 | 34 | 6239 | 0 | 1 | 0 | 0 |
| **Bare Soil** | 0 | 1 | 1 | 10 | 5 | 8726 | 8 | 10 | 0 |
| **Bitumen** | 0 | 0 | 0 | 1 | 0 | 121 | 6844 | 0 | 0 |
| **Bricks** | 0 | 0 | 0 | 2 | 0 | 0 | 0 | 40512 | 0 |
| **Shadows** | 0 | 0 | 0 | 0 | 0 | 0 | 0 | 4 | 2727 |

**Fig. 4.** Confusion matrix of the proposed method over PaviaC dataset

2. From Fig. 2, it is evident that the proposed method performs well even with very small training size. More specifically, it achieves 94.19%, 94.57%, 94.87% and 95.26% accuracy over PaviaU with 4.4%, 5%, 9% and 15% training data respectively. Similarly, the proposed method also achieves 98.46%, 98.52%, 98.63% and 98.7% accuracy over PaviaC with 4.4%, 5%, 9% and 15% training data respectively. The chart also shows that beyond these training sizes accuracy begins to saturate due to higher training samples.

In order to analyze the proposed model performance for class-wise accuracy, we depict confusion matrix for PaviaU and PaviaC datasets in Fig. 3 and Fig. 4 respectively. Highlighted diagonal elements represent accurately classified samples for each class whereas, non-diagonal elements represent error of omission and error of commission values. This quantitatively expresses the amount of agreement between the ground truth class and the predicted HS class.

The accuracy comparison of proposed model with existing state-of-the-art methods Lee et al. [18] and Hamida et al. [19] are shown in Table 5. Table 5 represent overall accuracy for PaviaU and PaviaC datasets respectively. These results are obtained from proposed model architecture after hundred epochs with 4.4% and 5% training size. From the report stated in Table 5, it can be deduced that proposed model is accurately able to classify minute features of the input HS images. It can also be observed that the proposed method performs better or equally well when compared with other state-of-the art methods. More specifically, the proposed method outperforms [18] and [19] by 9.58% and 5.61% on PaviaU dataset with 4.4% training data. Similarly, our model

**Table 5.** Performance comparison of the proposed method with existing approaches. All the results are computed with 7x7 spatial neighborhood as input to the network

|  | Proposed Method | | Hamida et al. [19] | | Lee et al. [18] | |
|---|---|---|---|---|---|---|
| *Train (%)* | 4.4% | 5% | 4.4% | 5% | 4.4% | 5% |
| *PaviaU* | **94.19** | **94.57** | 88.58 | 88.17 | 84.61 | 89.08 |
| *PaviaC* | **98.46** | **98.52** | 98.13 | 98.05 | 98.17 | 98.88 |

**Table 6.** Kappa coefficient value obtained for proposed method

| Dataset | Training size (%) | Kappa coefficient |
|---|---|---|
| PaviaU scene | 4.4 | 0.924 |
| | 5 | 0.929 |
| PaviaC scene | 4.4 | 0.978 |
| | 5 | 0.979 |

outperforms [18] and [19] by the margin of 5.49% and 6.4% on PaviaU dataset with 5% training data.

We also tabulate different values of kappa coefficient for the proposed model with training size of 4.4% and 5% in Table 6. Kappa coefficient is a discrete multivariate technique, suitable for analyzing remote sensing data due to its discrete and multivariate nature. It is measure of degree of enhancement by the classifier over pure random allocation of classes. As given in Table 6, the proposed method achieves robust kappa coefficient measures in both PaviaU and PaviaC dataset.

## 5    Conclusion

Hyperspectral image classification requires design and development of spatio-spectral feature aware CNN network. The proposed 3D CNN based architecture is successfully able to extract detailed spectral features from input HS image using 3D convolutional kernels in combination with localized residual connections. By using residual connections and robust 3D CNN design, we overcome the issue of dimensionality, overfitting of deeper layers and gradient vanishing while back-propagating. The proposed method outperforms recent state-of-the-art approaches for hyperspectral image classification on PaviaU and PaviaC datasets. The model still needs to be tested on low resolution hyperspectral images. In future, model robustness can be improved to effectively classify not only high-resolution image but also low-resolution images.